\pdfoutput=1
\documentclass{llncs}
\usepackage{multirow}
\usepackage{caption}
\usepackage{booktabs}
\usepackage{epsfig}
\usepackage{longtable}
\usepackage{rotating}
\graphicspath{{fig/}}

\begin{document}
\frontmatter          

\title{Deep CNNs for HEp-2 Cells Classification :\\
A Cross-specimen Analysis}
\author{Hongwei Li\inst{1} \and Jianguo Zhang\inst{2} \and Wei-Shi Zheng\inst{1}}

\authorrunning{Hongwei Li et al.} 

\institute{Sun Yat-sen University, Guangzhou, China
\email{hlli@dundee.ac.uk, wszheng@ieee.org} 
\and
CVIP, University of Dundee, United Kingdom\\
\email{j.n.zhang@dundee.ac.uk}}

\maketitle              

\begin{abstract}
Automatic classification of Human Epithelial Type-2 (HEp-2) cells staining patterns is an
important yet challenging problem in medicine.
Although both shallow and deep classification systems have been proposed,
the study of deep convolutional neural networks (CNNs) on this task is shallow to date.
In this paper, a deep convolutional architecture was proposed to address the classification problem under cross-specimen evaluation.
Then the effect of different data augmentation strategies was investigated on this task.
Extra training data were generated from specimen images using segmentation mask and ground true bounding box.
We also compared our deep framework with other methods including CNNs originally proposed for this task.

\keywords{Human Epithelial Type-2, Cell Classification, Deep Learning, Convolutional Architecture}
\end{abstract}
\section{Introduction}
Indirect Immunofluorescence (IIF) is a commonly-used methodology to identify the presence of Anti-Nuclear Antibody (ANA)
which facilitates the diagnosis of various autoimmune diseases.
Until now, this diagnostic process is performed by specialist's observing under a fluorescence microscope
and it relies heavily on the experience and expertise of the physicians.
Manual identification of these staining patterns suffers from intrinsic limitations related to visual evaluation by humans~\cite{foggia2010early}.
Thus, automatic pattern classification by computer vision techniques has been increasingly demanded.
During the past four years, various computer-assisted diagnosis(CAD) systems have been designed with image analysis
techniques in several classification challenges held by ICPR and ICIP~\cite{foggia2013benchmarking,foggia2014pattern,lovell2014performance}.
Current research in this topic is in its early age and there is still room for improvement, especially for deep learning methods.

 \begin{figure}[t]
\begin{center}
  \includegraphics[width=0.95\linewidth,height=0.35\linewidth]{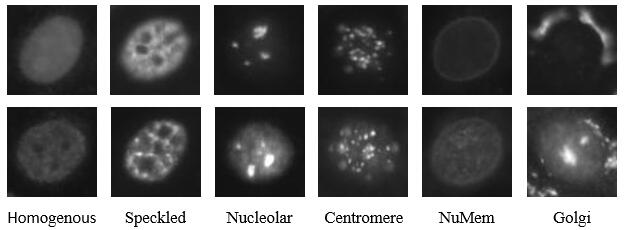}
\end{center}
   \caption{Sample cells images of six patterns from different specimens.}
\label{fig:figure1} 
\end{figure}

Due to the variations between patients and the photo-bleaching effect caused by the light source~\cite{song1995photobleaching},
the variations between separated specimens in same class/pattern are often large. As shown in Figure~\ref{fig:figure1}, some patterns could be easily misidentified (i.e., \emph{homogeneous} vs \emph{speckled}), indicating the inter-class variations on certain occasions are small.
The existing \textit{shallow} methods to address this problem mainly focus on three main separated aspects: handcrafting features, (sparse) coding and classification. Each of those components has been well studied; for instance, the winner of ICPR 2014 Classification Contest~\cite{manivannan2014hep} utilized multiple types of local descriptors with multi-resolution combining with sparse coding and ensemble SVMs. 
For these systems, hand-crafted features in feature-extraction stage, as well as parameter selection in feature-coding stage, rely much on empirical selection.

Very recently, deep learning methods, such as CNNs~\cite{krizhevsky2012imagenet}, have been initially applied to HEp-2 cells classification~\cite{gao2015hep,bayramoglu2015human}.
However, there are some key issues that have not been investigated:
i) cross-specimen evaluation of CNNs model and ii) the key factors in adopting deep CNNs for cell image classification.
It is important to note that the first issue is crucial to the evaluation of systems expected to be robust to different patients or specimens.
Thus it is particularly valuable for HEp-2 cells computer-assisted diagnosis system.

In this paper, we propose a new deep convolutional architecture and evaluated our system on public dataset released by ICPR~2014~Contest~\cite{hobson2013competition}.
Experimental result shows the our framework is effective and slightly outperforms another CNN-based system.
More importantly, we investigate further into the effect of different compositions of training set on deep model.
We find that data augmentation by adding new specimens are much more beneficial to overall performance than by employing affine transformations alone, which has been investigated in~\cite{gao2015hep,bayramoglu2015human}.

\section{Classification Framework}
\subsection{Datasets and Experimental Protocols}

 \begin{figure}[t]
\begin{center}
  \includegraphics[width=\linewidth,height=0.37\linewidth]{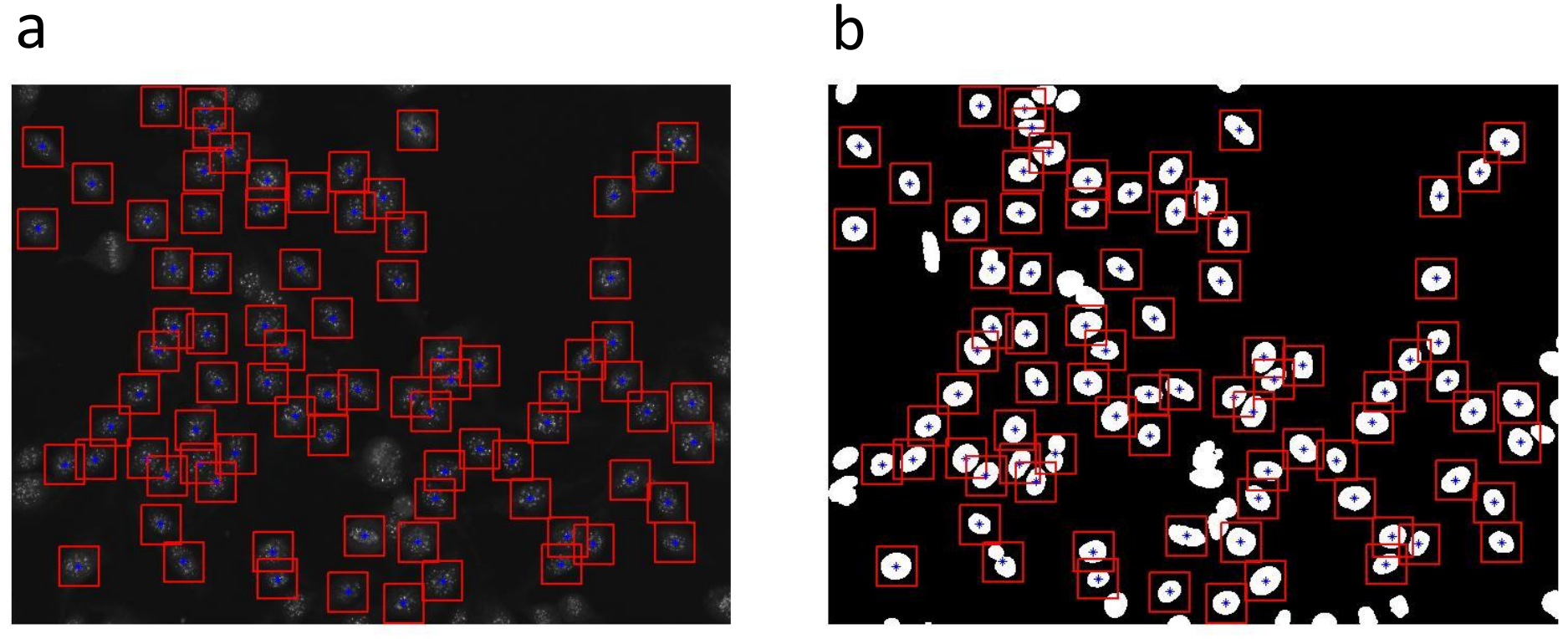}
\end{center}
   \caption{Extracting cells automatically from Task-2 dataset using ground truth bounding box ($77\times77$).
   (a) A centromere specimen; (b) The segmentation mask for specimen in (a).}
\label{fig:figure2} 
\end{figure}

\begin{table}{}
\vspace{0.01cm}
\scriptsize
\newcommand{\tabincell}[2]{\begin{tabular}{@{}#1@{}}#2\end{tabular}}
\renewcommand\arraystretch{1}
  \centering
\begin{tabular}{lccc}
\toprule

  \textbf{Patterns}&~~~\textbf{Task-1}&~~~~\textbf{Task-2}&~~~\textbf{Task-2 augmented}\\
\midrule
\textbf{Homogenous}&~~~2494&~~~~11386&~~~22772\\
\textbf{Speckled}&~~~2831&~~~~11858&~~~23716\\
\textbf{Nucleolar}&~~~2598&~~~~9320&~~~18640\\
\textbf{Centromere}&~~~2741&~~~~10199&~~~20398\\
\textbf{NuMem}&~~~2208&~~~~4363&~~~17452\\
\textbf{Golgi}&~~~724&~~~~1501&~~~12008
\\
\midrule
\textbf{Total}&~~~13596&~~~48627&~~~114986\\
\bottomrule
   \end{tabular}
   \vspace{0.1cm}
   \caption{Detailed compositions of Task-1, Task-2 and augmented Task-2 dataset}\label{table:Table1}
\end{table}

Two public available dataset: \textbf{Task-1} and \textbf{Task-2} from I3A Contest\cite{hobson2013competition} are used in our experiments. Task-1 dataset contains 13,596 cell images extracted from 83 specimens, with cells categorised  into six cell patterns namely \emph{Homogeneous},
\emph{Speckled}, \emph{Nucleolar}, \emph{Centromere}, \emph{Nuclear Membrane}, and \emph{Golgi}.
Task-2 dataset contains 237 specimens images with the six patterns above.
As shown in Fig.~\ref{fig:figure2}, a simple automatic procedure is employed to segment cells from each specimen given the segmentation mask in the dataset.
48,627 cell images (roughly 200 cells from each specimen) are obtained after human selection from Task-2 dataset and are then used as an extra training set for deep model.
For comparative study, we employ data augmentation on Task-2.
Specifically, the number of images from first four patterns are doubled by rotating 90 degrees whilst images from the last two patterns named \emph{NuMem} and \emph{Golgi} are rotated by 90, 180, 270 degrees and mirrored respectively to generate eight times larger data, considering the last two patterns contain relatively less images.

Leave-one-specimen-out (LOSO) is an important setting to test the generalisation performance of a computer-assisted diagnosis system~\cite{qi2015hep,manivannan2016automated}. In this paper, we adopt the setting in our evaluation.
In the LOSO strategy, each time all cell images from one of the 83 specimens are used for testing, the rest are used for training. Mean-class-accuracy (MCA) is used as evaluation metric based on the 83 splits, which is defined as
\begin{equation}
 MCA = \frac{1}{K}\sum _{k = 1}^{K}CCR_{k}
\end{equation}
where \small{$CCR_{k}$} is the correct classification rate for class $k$ and $K$ is the total number of classes.
This metric was required by the I3A HEp-2 classification contest.

\subsection{Deep Convolutional Architecture}

\begin{figure}[t]
\begin{center}
  \includegraphics[width=\linewidth,height=0.3\linewidth]{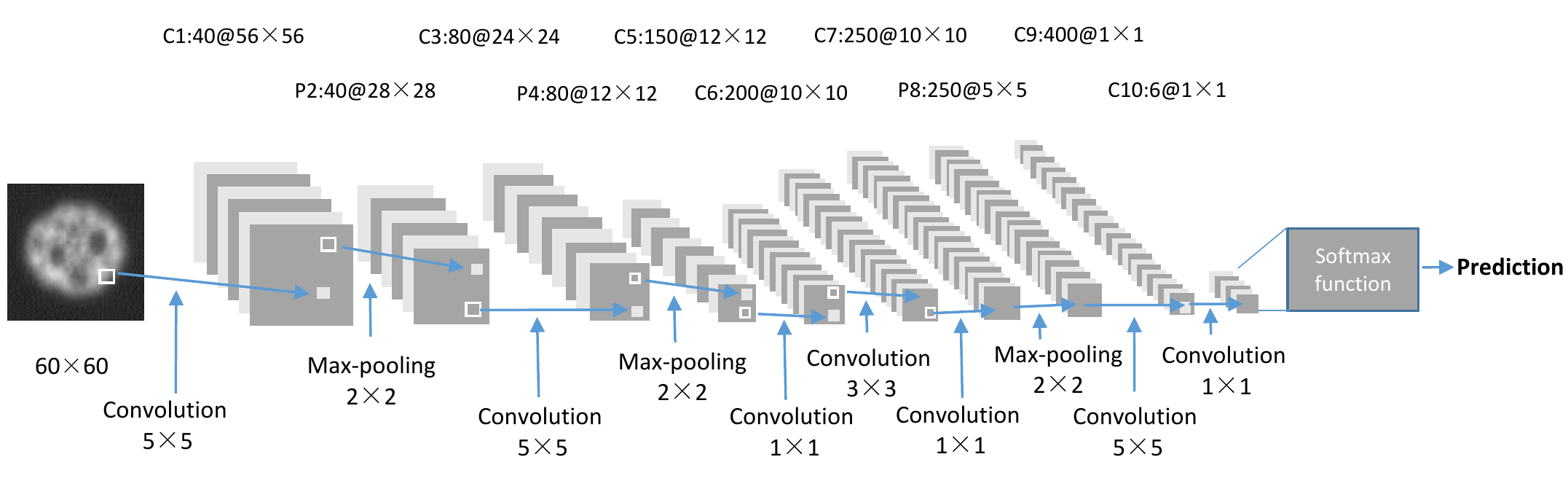}
\end{center}
   \caption{The architecture of our deep CNNs.}
\label{fig:figure3} 
\end{figure}

Our deep CNNs is inspired by the architecture of the recent works\cite{43022,lin2013network}.
Specifically, it contains ten layers as illustrated in Figure~\ref{fig:figure3} which shares the basic structure of CNNs.
But different from the classic CNNs - LeNet \cite{lecun1998gradient}, convolutional layers with $1\times1$ kernel size are heavily used in our model to increase the depth and number of weights.
The network is initialized with small random numbers (around 0.001).
Rectified linear units (ReLU) are employed as activation functions after convolutional layers.
The model is trained with batch size of 200 and 50 epoches, and the learning rate is set to $0.002$.
These settings are deliberately kept fixed for all experiments to observe the effect of changes in the training data.
In this work, we use MatConvNet~\cite{vedaldi2015matconvnet} library to build up deep convolutional architecture.
Usually, normalization step is not an essential part of CNNs as the networks have capacity to handle variations.
Therefore, we did not implement normalization on the data.
Each image was resized to $60\times60$ to guarantee a uniform scale of all the images used for training.
We augment the training set in two ways: 1) by affine transformations, e.g. rotations; 2) by importing extra data from Task-2 dataset.


To classify a test image, firstly the image is resized to $60\times60$. Then the image is forward-propagated through
the network, and the expected score for each class is obtained.
To further improve the robustness of our system, we select the last three epoches, that is, the 48th, 49th, and 50th epoch to jointly classify a test image.
The predicted label is the one with the maximum output score averaged over the three votes.
Although fine-tuning has been proven successful on a variety of classification tasks,
the experimental results show that it achieves poor performance in this task by using the pre-trained model on ImageNet.


%
%
%
%

\section{Comparative Study on Data Augmentation}
Data augmentation is a key step for training CNN when data is limited and it is also the case of recent study of using CNN for HEp-2 cell classification. However, only affine transformations for augmentation was ever considered in this problem. As one can observe from each specimen, cells inside have little variations in texture but positions and orientations, thus employing random rotations and mirroring has limited improvements on training the deep model.
In this section we investigate the effect of augmentation when adding extra specimens. Then experimental results show that our CNNs is effective under different augmentation strategies.

In total, three groups of experiments are designed to explore the effects of additional specimens under \emph{leave-one-specimen-out} evaluation: 1) \textit{set-1}: only Task-1 dataset is used; 2) \textit{set-2}: bigger dataset obtained by data augmentation on Task-1 only; 3) \textit{set-3}: dataset combining Task-1 with additional 237 specimens from Task-2.
In building \textit{set-2}, each image in Task-1 is rotated by angle $0^{0}$, $90^{0}$, $180^{0}$ and $270^{0}$ and mirrored respectively to generate new images. The size of \textit{set-2} is eight times larger than Task-1.
In building \textit{set-3}, we use Task-1 dataset and additional 237 specimens from Task-2.
For each specimens, 41 cells images are randomly selected within each specimen.
Thus we obtain a dataset with size similar to Task-1 but with more specimens contained.

\begin{table}{}
\vspace{0.01cm}
\scriptsize
\newcommand{\tabincell}[2]{\begin{tabular}{@{}#1@{}}#2\end{tabular}}
\renewcommand\arraystretch{1}
  \centering
\begin{tabular}{lllllllc}
\toprule

  \textbf{Patterns}&~~~\textbf{\textit{set-1}}&~~~\textbf{\textit{set-2}}&~~~\textbf{\textit{set-3}}\\
\midrule
\textbf{Homogenous}&~~~2494&~~~19952&~~~2829\\
\textbf{Speckled}&~~~2831&~~~22648&~~~2788\\
\textbf{Nucleolar}&~~~2598&~~~20784&~~~2706\\
\textbf{Centromere}&~~~2741&~~~21928&~~~2747\\
\textbf{NuMem}&~~~2208&~~~17664&~~~1476\\
\textbf{Golgi}&~~~724&~~~5792&~~~574\\
\midrule
\textbf{Total}&~~~13596&~~~108768&~~~13120\\
\bottomrule
   \end{tabular}
   \vspace{0.1cm}
   \caption{Detailed compositions of data set in three groups of experiments}\label{table:Table5}
\end{table}

Table~\ref{table:Table6} shows the experimental results and Table~\ref{table:Table7} shows the confusion matrix of result on \textit{set-3}. By comparing the results on \textit{set-1} and \textit{set-2}, we see that model trained with data augmentation outperforms the former one by $4.15\%$,
demonstrating that data augmentation is necessary and effective for small training sets in CNNs frameworks.
Notably, the model trained on \textit{set-3} with less images but more specimens than \textit{set-1},
outperforms one trained on \textit{set-2}, which uses eight times larger dataset.
It indicates that data augmentation by adding new specimens are more powerful and more efficient than by employing rotations and mirroring alone.

\begin{table}{}
\vspace{0.1cm}
\scriptsize
\newcommand{\tabincell}[2]{\begin{tabular}{@{}#1@{}}#2\end{tabular}}
\renewcommand\arraystretch{1}
  \centering
\begin{tabular}{lc}
\toprule

  \textbf{Training Set}~~~&\textbf{Mean Class Accuracy}\\
\midrule
\textbf{Set-1}~~~&\textbf{70.52$\%$} \\

\textbf{Set-2}~~~&\textbf{74.67$\%$} \\

\textbf{Set-3}~~~&\textbf{79.13$\%$} \\
\bottomrule
   \end{tabular}
   \vspace{0.1cm}
   \caption{Comparisons of performance on different training sets}\label{table:Table6}
\end{table}
\begin{table}{}
\vspace{0.01cm}
\scriptsize
\newcommand{\tabincell}[2]{\begin{tabular}{@{}#1@{}}#2\end{tabular}}
\renewcommand\arraystretch{1}
  \centering
\begin{tabular}{lllllllc}
\toprule

  \textbf{Patterns}&~~~\textbf{Homogenous}&\textbf{Speckled}&\textbf{Nucleolar}&\textbf{Centromere}&\textbf{NuMem}&\textbf{Golgi}\\
\midrule
\textbf{Homogenous}&~~~\textbf{86.16}&~~~9.06&~~~0.56&~~~0&~~~4.21&~~~0\\
\textbf{Speckled}  &~~~12.86&~~~\textbf{70.96}&~~~8.72&~~~5.30&~~~2.01&~~~0.14\\
\textbf{Nucleolar} &~~~2.61&~~~3.27&~~~\textbf{86.87}&~~~3.23&~~~2.54&~~~1.46\\
\textbf{Centromere}&~~~0&~~~10.94&~~~4.85&~~~\textbf{83.62}&~~~0.36&~~~0.22\\
\textbf{NuMem}     &~~~5.48&~~~3.13&~~~0.68&~~~0.14&~~~\textbf{85.42}&~~~5.16\\
\textbf{Golgi}     &~~~11.88&~~~4.14&~~~7.73&~~~1.10&~~~13.39&~~~\textbf{61.74}\\
\bottomrule

   \end{tabular}
   \vspace{0.1cm}
   \caption{Confusion matrix of \textit{set-3}. The mean class accuracy obtained is 79.13\%.}\label{table:Table7}
\end{table}

In previous CNN-based framework\cite{gao2015hep}, data augmentation was based on affine transformations.
 It was shown effective under \emph{\textbf{k-fold cross-validation}} protocol.
However, under \emph{k-fold cross-validation} protocol,
the identities of specimen from which cells were extracted were totally disregarded.
As one can observe from each specimen, the cells inside have little variation in texture but positions and orientations, which causes that the training set and test set are highly similar. Thus the task will be artificially made easier in this setting.
The mean-class-accuracy of our CNNs under this setting is 91.16\% without data augmentation.
Therefore, one system achieving high recognition accuracy under \emph{k-fold cross-validation}  may achieve relatively poor performance under \emph{cross-specimen} test.

Further experiments are conducted based on the datasets in Section $2.1$ to demonstrate the effectiveness.
A pooled set of 48,627 images from Task-2 and training set of Task-1 were used for training.
Then a mean-class-accuracy of 82.90\% was obtained.
We further employed data augmentation on Task-2 by rotation illustrated in Section 2.1.
By training with a pooled set of 114,986 images and training set of Task-1, a slightly better classification accuracy of 83.55\% was obtained.
Table~\ref{table:Table3} reported the confusion matrix. The Golgi class had poor results (70.30\%).

\begin{table}{}
\vspace{0.01cm}
\scriptsize
\newcommand{\tabincell}[2]{\begin{tabular}{@{}#1@{}}#2\end{tabular}}
\renewcommand\arraystretch{1}
  \centering
\begin{tabular}{lllllllc}
\toprule

  \textbf{Patterns}&~~~\textbf{Homogenous}&\textbf{Speckled}&\textbf{Nucleolar}&\textbf{Centromere}&\textbf{NuMem}&\textbf{Golgi}\\
\midrule
\textbf{Homogenous}&~~~\textbf{85.93}&~~~8.02&~~~0.20&~~~0&~~~3.65&~~~2.21\\
\textbf{Speckled}  &~~~9.71&~~~\textbf{79.97}&~~~3.50&~~~5.58&~~~1.24&~~~0\\
\textbf{Nucleolar} &~~~2.19&~~~3.66&~~~\textbf{85.84}&~~~6.66&~~~0.92&~~~0.73\\
\textbf{Centromere}&~~~0.04&~~~10.58&~~~4.16&~~~\textbf{84.93}&~~~0.11&~~~0.18\\
\textbf{NuMem}     &~~~1.45&~~~1.18&~~~0.68&~~~0.09&~~~\textbf{94.34}&~~~2.26\\
\textbf{Golgi}     &~~~5.11&~~~5.52&~~~13.54&~~~0.97&~~~4.56&~~~\textbf{70.30}\\
\bottomrule

   \end{tabular}
   \vspace{0.1cm}
   \caption{Confusion matrix on \textit{Task-1} under \emph{leave-one-specimen-out} protocol.
     The MCA is 83.55\%.}\label{table:Table3}
\end{table}

\section{Comparison with State-of-the-Art}

%
We further compare our approach with several state-of-the-art systems: 1)~CNNs which shares the basic structure of LeNet-5~\cite{gao2015hep}; 2)~Shape index histograms with donut-shaped spatial pooling~\cite{larsen2014hep}; 3) multi-resolution local descriptor with ensemble SVMs (the winner of ICPR14 contest)~\cite{manivannan2016automated}.
These methods were top performers on Task-1 dataset.
Since these methods employ only Task-1 dataset, we choose the experimental result on \textit{Set-2} for fair comparison.

%
%

\begin{table}{}
\vspace{0.1cm}
\scriptsize
\newcommand{\tabincell}[2]{\begin{tabular}{@{}#1@{}}#2\end{tabular}}
\renewcommand\arraystretch{1}
  \centering
\begin{tabular}{lc}
\toprule

  \textbf{Methods}&\textbf{Mean Class Accuracy}\\
\midrule
CNNs shares the basic structure of LeNet-5~\cite{gao2015hep}&71.88$\%$ \\
\textbf{Our proposed CNNs}&\textbf{74.67$\%$} \\

Shape index histograms with donut-shaped spatial pooling~\cite{larsen2014hep}&78.70$\%$ \\

Multi-resolution patterns with Ensemble SVMs~\cite{manivannan2016automated} &81.10$\%$ \\


\bottomrule
   \end{tabular}
   \vspace{0.1cm}
   \caption{Comparisons with state-of-the-art }\label{table:Table4}
\end{table}

As shown in Table~\ref{table:Table4}, our proposed method slightly outperforms the former CNNs by 2.79\% demonstrating the effectiveness of deeper model. However, the overall performance of CNNs remains poor compared with traditional methods. Since cells from one specimen have little variations in texture but positions and orientations, the training set containing only 82 specimens could be considered rather small. In Section 3, our model achieves accuracy of 83.55\% when training with extra specimens, demonstrating deep model holds potential on this task.

\section{Conclusions}

In this work, we presented a detailed study of a deep convolutional architecture to address HEp-2 cell classification problem.
We evaluated our systems in cross-specimen experiments.
In particular, we investigated the effect of different data augmentation strategies.
Our results show that the key factor for training a good deep model is by adding sufficient specimens and affine transformation, which could calibrate further development of using CNN in classifying HEp-2 cell images.

{
\bibliographystyle{splncs03}
\bibliography{egbib}

\begin{thebibliography}{10}
\providecommand{\url}[1]{\texttt{#1}}
\providecommand{\urlprefix}{URL }

\bibitem{bayramoglu2015human}
Bayramoglu, N., Kannala, J., Heikkila, J.: Human epithelial type 2 cell
  classification with convolutional neural networks. In: Bioinformatics and
  Bioengineering (BIBE), 2015 IEEE 15th International Conference on. pp. 1--6.
  IEEE (2015)

\bibitem{foggia2014pattern}
Foggia, P., Percannella, G., Saggese, A., Vento, M.: Pattern recognition in
  stained hep-2 cells: Where are we now? Pattern Recognition  47(7),
  2305--2314 (2014)

\bibitem{foggia2010early}
Foggia, P., Percannella, G., Soda, P., Vento, M.: Early experiences in mitotic
  cells recognition on hep-2 slides. In: Computer-based medical systems (CBMS),
  2010 IEEE 23rd international symposium on. pp. 38--43. IEEE (2010)

\bibitem{foggia2013benchmarking}
Foggia, P., Percannella, G., Soda, P., Vento, M.: Benchmarking hep-2 cells
  classification methods. Medical Imaging, IEEE Transactions on  32(10),
  1878--1889 (2013)

\bibitem{gao2015hep}
Gao, Z., Wang, L., Zhou, L., Zhang, J.: Hep-2 cell image classification with
  deep convolutional neural networks. arXiv preprint arXiv:1504.02531  (2015)

\bibitem{hobson2013competition}
Hobson, P., Percannella, G., Vento, M., Wiliem, A.: Competition on cells
  classification by fluorescent image analysis. In: Proc. 20th IEEE Int. Conf.
  Image Process.(ICIP). pp. 2--9 (2013)

\bibitem{krizhevsky2012imagenet}
Krizhevsky, A., Sutskever, I., Hinton, G.E.: Imagenet classification with deep
  convolutional neural networks. In: Advances in neural information processing
  systems. pp. 1097--1105 (2012)

\bibitem{larsen2014hep}
Larsen, A.B.L., Vestergaard, J.S., Larsen, R.: Hep-2 cell classification using
  shape index histograms with donut-shaped spatial pooling. Medical Imaging,
  IEEE Transactions on  33(7),  1573--1580 (2014)

\bibitem{lecun1998gradient}
LeCun, Y., Bottou, L., Bengio, Y., Haffner, P.: Gradient-based learning applied
  to document recognition. Proceedings of the IEEE  86(11),  2278--2324 (1998)

\bibitem{lin2013network}
Lin, M., Chen, Q., Yan, S.: Network in network. arXiv preprint arXiv:1312.4400
  (2013)

\bibitem{lovell2014performance}
Lovell, B.C., Percannella, G., Vento, M., Wiliem, A.: Performance evaluation of
  indirect immunofluorescence image analysis systems. ICPR 2014  (2014)

\bibitem{manivannan2014hep}
Manivannan, S., Li, W., Akbar, S., Wang, R., Zhang, J., McKenna, S.J.: Hep-2
  cell classification using multi-resolution local patterns and ensemble svms.
  In: Pattern Recognition Techniques for Indirect Immunofluorescence Images
  (I3A), 2014 1st Workshop on. pp. 37--40. IEEE (2014)

\bibitem{manivannan2016automated}
Manivannan, S., Li, W., Akbar, S., Wang, R., Zhang, J., McKenna, S.J.: An
  automated pattern recognition system for classifying indirect
  immunofluorescence images of hep-2 cells and specimens. Pattern Recognition
  51,  12--26 (2016)

\bibitem{qi2015hep}
Qi, X., Zhao, G., Chen, J., Pietik{\"a}inen, M.: Hep-2 cell classification: The
  role of gaussian scale space theory as a pre-processing approach. Pattern
  Recognition Letters  (2015)

\bibitem{song1995photobleaching}
Song, L., Hennink, E., Young, I.T., Tanke, H.J.: Photobleaching kinetics of
  fluorescein in quantitative fluorescence microscopy. Biophysical journal
  68(6),  2588 (1995)

\bibitem{43022}
Szegedy, C., Liu, W., Jia, Y., Sermanet, P., Reed, S., Anguelov, D., Erhan, D.,
  Vanhoucke, V., Rabinovich, A.: Going deeper with convolutions. In: CVPR 2015
  (2015), \url{http://arxiv.org/abs/1409.4842}

\bibitem{vedaldi2015matconvnet}
Vedaldi, A., Lenc, K.: Matconvnet: Convolutional neural networks for matlab.
  In: Proceedings of the 23rd Annual ACM Conference on Multimedia Conference.
  pp. 689--692. ACM (2015)

\end{thebibliography}
}

\end{document}